\documentclass[twocolumn]{article}
\usepackage{mglgrid}

\pdfoutput=1

\usepackage[utf8]{inputenc} % allow utf-8 input
\usepackage[T1]{fontenc}    % use 8-bit T1 fonts
\usepackage[hidelinks]{hyperref}
\usepackage{url}            % simple URL typesetting
\usepackage{booktabs}       % professional-quality tables
\usepackage{amsfonts}       % blackboard math symbols
\usepackage{nicefrac}       % compact symbols for 1/2, etc.
\usepackage{microtype}      % microtypography
\usepackage[title]{appendix}
\usepackage{pgfplots}
\pgfplotsset{compat = newest}
\usepackage{graphicx}
\usepackage{booktabs}
\usepackage{multirow}
\usepackage{xspace}
\usepackage{booktabs}
\usepackage{float}
\usepackage{color}
\usepackage{mathtools}
\usepackage{amsmath}
\usepackage{amsthm}
\usepackage{xcolor}
\usepackage[normalem]{ulem}
\usepackage{siunitx}
\usepackage{makecell}
\usepackage[]{algpseudocode}
\usepackage{algorithm}

\usepackage[font=small]{caption}
\usepackage{subcaption}
\usepackage{tabularx}
\usepackage{cleveref}

\DeclarePairedDelimiter{\floor}{\lfloor}{\rfloor}

\newtheorem*{assumption*}{\assumptionnumber}
\providecommand{\assumptionnumber}{}
\makeatletter

\newcommand{\vb}{\boldsymbol{b}}
\newcommand{\vx}{\boldsymbol{x}}
\newcommand{\vc}{\boldsymbol{c}}
\newcommand{\vh}{\boldsymbol{h}}
\newcommand{\vcprev}{\boldsymbol{c}_{\textit{prev}}}
\newcommand{\vhprev}{\boldsymbol{h}_{\textit{prev}}}
\newcommand{\vw}{\boldsymbol{w}}
\newcommand{\vi}{\boldsymbol{i}}
\newcommand{\vj}{\boldsymbol{j}}
\newcommand{\vf}{\boldsymbol{f}}
\newcommand{\vo}{\boldsymbol{o}}
\newcommand{\mW}{\mathbf{W}}
\newcommand{\mE}{\mathbf{E}}
\newcommand{\mM}{\mathbf{M}}
\newcommand{\mQ}{\mathbf{Q}}
\newcommand{\mR}{\mathbf{R}}

\DeclareMathOperator*{\LSTM}{LSTM}
\DeclareMathOperator*{\RLSTM}{RLSTM}
\DeclareMathOperator*{\mogrify}{mogrify}
\DeclareMathOperator*{\softmax}{softmax}
\DeclareMathOperator*{\onehot}{onehot}

\newcommand{\wikitexttwo}{Wikitext-2\xspace}
\newcommand{\enwik}{Enwik8\xspace}
\newcommand{\texteight}{Text8\xspace}
\newcommand{\nlltoppl}[1]{%
  \pgfmathparse{exp(#1)}%
  \pgfmathprintnumber[fixed,zerofill,precision=1,assume math mode=true]{\pgfmathresult}}
\newcommand{\nlltopplbold}[1]{%
  \pgfmathparse{exp(#1)}%
  \textbf{\pgfmathprintnumber[fixed,zerofill,precision=1,assume math mode=true]{\pgfmathresult}}}
\newcommand{\nlltobpc}[1]{%
  \pgfmathparse{log2(exp(#1))}%
  \pgfmathprintnumber[fixed,zerofill,precision=3,assume math mode=true]{\pgfmathresult}}
\newcommand{\nlltobpcbold}[1]{%
  \pgfmathparse{log2(exp(#1))}%
  \textbf{\pgfmathprintnumber[fixed,zerofill,precision=3,assume math mode=true]{\pgfmathresult}}}

\newcommand\nobreakpar{\par\nobreak\@afterheading}
\makeatother

\newcommand{\tta}{\textlf{2}TA}
\definecolor{mglred}{RGB}{163,65,48}
\newcommand{\tmid}{\hspace{0.08em}|\hspace{0.11em}}
\DeclarePairedDelimiterX{\condp}[2]{(}{)}{%
  #1\;\delimsize|\penalty 0 \;#2%
}

\allowdisplaybreaks

\title{Circling Back to Recurrent\\Models of Language}

\author{\textit{G\'abor Melis} \\
  {\tt melisgl@google.com}\\
  DeepMind, UCL; London, UK
}

\begin{document}

\maketitle

\begin{abstract}
Just because some purely recurrent models suffer from being hard to optimize and inefficient on today's hardware, they are not necessarily bad models of language.
We demonstrate this by the extent to which these models can still be improved by a combination of a slightly better recurrent cell, architecture, objective, as well as optimization.
In the process, we establish a new state of the art for language modelling on small datasets and on \enwik with dynamic evaluation.
\end{abstract}

\section{Introduction}

Reliable model comparison is crucial for continued innovation in language modelling.
With so much attention on Transformers \citep{vaswani2017attention}, the development of purely recurrent models has ebbed away.
It might be tempting to claim that purely recurrent models are fundamentally worse models of language than attention-based ones, but they seem to have an edge on small datasets, while on larger datasets their lack of scalability on current hardware \citep{DBLP:journals/corr/abs-2009-06489} also plays an important role in their evaluation.
Due to being unfashionable and slow, their fitness might be underestimated especially on all but the smallest datasets with the exception of S4 \citep{gu2021efficiently}, a highly parallelizable, long-range model.
The purpose of this work is simply to apply more resources to advancing and evaluating recurrent models --~despite, and to compensate for, their computational inefficiency~-- to better represent their performance in future model comparisons.

Starting from the Mogrifier LSTM \citep{melis2019mogrifier}, a state-of-the-art recurrent language model, we introduce several small changes, which involve not only the design of the recurrent cell but the overall architecture, the training objective, and optimization.
On all datasets, they combine to a significant, and in some cases large, effect.

\section{Rewired LSTM}

We propose a novel recurrent cell, a slightly tweaked version of the LSTM \citep{hochreiter1997lstm}.
In an LSTM, the updated state $\vc$ and the output $\vh$ are computed as
{\setlength{\jot}{0.5em}
\begin{align*}
  \vi &= \sigma\bigl(\mW^{ix} \vx + \mW^{ih} \vhprev + \vb^i\bigr)\\
  \vj &= \tanh(\mW^{jx}\vx + \mW^{jh} \vhprev + \vb^j) \\
  \vf &= \sigma(\mW^{fx} \vx + \mW^{fh} \vhprev + \vb^f) \\
  \vc &= \vf \odot \vcprev + \vi \odot \vj \\
  \vo &= \sigma(\mW^{ox} \vx + \mW^{oh} \vhprev + \vb^o) \\
  \vh &= \vo \odot \tanh(\vc),
\end{align*}}%
where $\sigma$ is the logistic sigmoid function, $\odot$ is the elementwise product, $\mW^{**}$ and $\vb^{*}$ are weight matrices and biases.
With $m$ and $n$ being the sizes of the input and the cell state, let the $\LSTM \colon \mathbb{R}^n\times\mathbb{R}^n \times \mathbb{R}^m \to \mathbb{R}^n \times \mathbb{R}^n$ function refer to the cell computation given above: $\LSTM(\vcprev, \vhprev, \vx) = (\vc, \vh)$.
To accommodate the performance characteristics of parallel hardware, the activations of $\vi$, $\vj$, $\vf$, $\vo$ are in practice often computed by tiling the eight $\mW^{**}$ into one large matrix and multiplying it with the concatenated input and recurrent state $[\vx, \vhprev]$.

Intuitively, in $\vc = \vf \odot \vcprev + \vi \odot \vj$ the forget gate $\vf$ and the proposed update $\vi \odot \vj$ depend on each other, and computing them in parallel from the same inputs may be partially redundant.
On the flipside, the parametrization of the proposed update as a product is more expressive than that of the forget gate, so the overall cell update may lose some of the extra expressivity.
Hence, it makes sense to compute the forget gate from the proposed update instead.
Note that this reduces the opportunities for parallelization and makes cell updates slower.

Further disregarding efficiency on today's hardware, to reduce parameter count and to encourage storing more information in the cell state $\mathbf{c}$, we compute $\vo$ from $\mathbf{c}$, dropping a potential bypass connection from $\vx$ to $\vh$.
Finally, to make exploding gradients less likely, we cap $\vi$ at $\mathbf{1-f}$ to ensure $|c_u| \leqslant 1$ for all memory units $u$.
The update of our Rewired LSTM (RLSTM) cell takes the form
{\setlength{\jot}{0.5em}
\begin{align*}
  \vi &= \sigma\bigl(\mW^{ix} \vx + \mW^{ih} \vhprev + \vb^i\bigr) \\
  \vj &= \tanh\bigl(\mW^{jx}\vx + \mW^{jh} \vhprev + \vb^j\bigr) \\
  \vf &= \sigma\bigl(\mW^{fu} \textcolor{mglred}{i\odot j} + \mW^{fh} \vhprev + \vb^f\bigr) \\
  \vc &= \vf \odot \vcprev + \textcolor{mglred}{\min(\vi, 1-\vf)} \odot \vj \\
  \vo &= \sigma\bigl(\textcolor{mglred}{\mW^{oc} \vc} + \vb^o\bigr) \\
  \vh &= \vo \odot \tanh\bigl(\vc\bigr),
\end{align*}}%
where the changes from the LSTM are highlighted in red.
Based on this altered computation, we define the $\RLSTM$ function similarly to the $\LSTM$ above.

\section{Architecture}
\label{sec:architecture}

The way recurrent cells are combined can also be improved.
Here, we opt to use residual connections, where previous works have used stacked LSTMs \citep{merity2017regularizing} or skip connections \citep{melis2019mogrifier}, which feed directly into the final output.
Along with this change, we apply dropout \citep{hinton2012improving} to the cell output before it is added to the residual branch.

To describe the overall architecture more formally, let us denote the mogrification operation \citep{melis2019mogrifier} with the following $\mathbb{R}^n\times\mathbb{R}^m \to \mathbb{R}^n \times \mathbb{R}^m$ function:
\begin{align*}
  \mogrify(\vh,\vx) &= \vh^{2\floor{r/2}}, \vx^{2\floor{(r+1)/2}-1}\\
  \vx^{-1}, \vh^0 &= \vx, \vh
\end{align*}
\begin{align*}
  \vx^i &= 2\sigma\bigl(\mQ^{i}\vh^{i-1}\bigr) \odot \vx^{i-2} & \text{for odd i} \in [1..r],\\
  \vh^i &= 2\sigma\bigl(\mR^{i}\vx^{i-1}\bigr) \odot \vh^{i-2} & \text{for even i} \in [1..r],
\end{align*}
where the number of rounds $r$ is a hyperparameter, and $\mQ^{i} \in \mathbb{R}^{m\times n}, \mR^{i} \in \mathbb{R}^{n\times m}$ are (possibly low-rank factorized) weight matrices.

Denoting time steps with $t \in [1, 2, \dots]$ and layers $l \in [1..L]$, from the vector of token indices $\vw$  in a fixed vocabulary, the probability distribution of the next token $p(.\tmid \vw_{<t},\mM_t)$ is computed as
{\setlength{\jot}{0.6em}
\begin{align*}
\vc^l_0, \vh^l_0 &= \mathbf{0}, \mathbf{0}\\
% Kerning is funny, probably due to lining figures. Add some negative kerns.
\hat{\vx}^0_t\! &= \onehot\bigl(\vw_t\bigr) \mE^{\textrm{in}} \odot \mM^\textrm{in}_t\\
\hat{\vx}^l_t &= \vh^l_t \odot \mM^{\textrm{cell},l}_t \qquad \qquad (l > 1) \\
\hat{\vh}^l_t &= \vh^l_t \odot \mM^{\textrm{state},l}\\
\vc^1_t\!, \vh^1_t\! &= \textrm{[R]LSTM}\Bigl(\vc^0_{t-1}, \mogrify\Bigl(\hat{\vh}^1_{t-1}, \hat{\vx}^0_t\Bigr)\Bigr)
\end{align*}
\begin{align*}
\vc^l_t, \vh^l_t &= \textrm{[R]LSTM}\Bigl(\vc^l_{t-1}, \mogrify\Bigl(\hat{\vh}^l_{t-1}, \sum\nolimits_{i=1}^{l-1} \hat{\vx}^i_t\Bigr)\Bigr)\\
p(.\tmid \vw_{<t},\mM_t) &= \softmax\Bigl(\Bigl(\sum\nolimits_{l=1}^L \hat{\vx}^l_t\Bigr) \odot \mM^{\textrm{out}}_t \mE^{\textrm{out}} + \vb^{\textrm{out}}\Bigr),
\end{align*}}%
where we assume that $m=n$ to allow for a residual architecture without projections, $\mE^{\textrm{in}}$ and $\mE^{\textrm{out}}$ are the input and output embedding matrices.
For word-based language modelling, we set $\mE^{\textrm{out}}$ to the transpose of $\mE^{\textrm{in}}$ \citep{DBLP:journals/corr/ZophL16,DBLP:journals/corr/PressW16}.
$\mM_t$ is the set of individual input, cell, state and output dropout mask matrices $\smash{\mM^{\textrm{in}}_t}$, $\smash{\mM^{\textrm{cell},l}_t}$, $\smash{\mM^{\textrm{state},l}}$ and $\smash{\mM^{\textrm{out}}_t}$.
Note that for state dropout, we use the variational dropout of \citet{gal2016theoretically}, thus $\mM^{\textrm{state},l}$ does not depend on $t$.
In addition, when an RLSTM is used instead of an LSTM cell, we also apply the state dropout mask to $\vc$ in the calculation of $\vo = \sigma(\mW^{oc} (\vc \odot \mM^{\textrm{state},l}) + \vb^o)$.% of the corresponding layers.

\section{Objective}

In the objective, we average model predictions over multiple dropout samples:
\begin{align*}
\ln p\condp[\big]{\vw_t}{\vw_{<t}} = \ln\biggl(\frac{1}{D} \sum_{d=1}^D p\condp[\big]{\vw_t}{\vw_{<t},\mM^d_t}\biggl),
\end{align*}
where $D$ is the number of samples taken.
As pointed out by \citet{noh2017regularizing}, when dropout is interpreted as optimizing a variational lower bound on the log likelihood \citep{gal2016theoretically}, this procedure is an instantiation of Importance Weighted Autoencoders \citep{burda2015importance}, which provide a bound tighter than the single-sample ELBO.

\begin{table*}
  \setlength{\abovecaptionskip}{0.2\baselineskip}
  \caption[Word-level perplexities of near state-of-the-art models.]{Word-level perplexities of near state-of-the-art models.
Names for models with our new results are in bold.
On \wikitexttwo, the baseline employed Mixture of Softmaxes \citep{yang2017improved}, but we found no benefit to that when used in conjunction with multiple dropout samples.}
  \label{tab:cb-word-results}

  \centering
  \small
  \begin{tabular}{@{}llrlrlr@{}}
    & & & \multicolumn{2}{c}{No Dyneval} & \multicolumn{2}{c}{Dyneval} \\
    \cmidrule(lr){4-5} \cmidrule(l){6-7}
    & & & Val. & Test & Val. & Test \\
    \midrule
    \parbox[t]{5mm}{\multirow{4}{*}{\rotatebox[origin=c]{90}{PTB}}}
    & Transformer-XL \citep{dai2019transformer} & 24M
        & 56.7 & 54.5 & & \\
    & Mogrifier LSTM \citep{melis2019mogrifier} & 24M
        & \nlltoppl{3.95346} & \nlltoppl{3.93124}
        & \nlltoppl{3.80963} & \nlltoppl{3.80708} \\
    & \textbf{Mogrifier LSTM} & 24M
        & \nlltoppl{3.91065} & \nlltoppl{3.88170}
        & \nlltoppl{3.77302} & \nlltoppl{3.76763} \\
    & \textbf{Mogrifier RLSTM} & 24M
        & \nlltoppl{3.88915} & \nlltopplbold{3.86939}
        & \nlltoppl{3.75818} & \nlltopplbold{3.75805} \\
    \midrule
    \parbox[t]{5mm}{\multirow{3}{*}{\rotatebox[origin=c]{90}{WT$2$}}}
    & Mogrifier LSTM MoS$2$ & 35M
        & \nlltoppl{4.07235} & \nlltoppl{4.03665}
        & \nlltoppl{3.70429} & \nlltoppl{3.66475} \\
    & \textbf{Mogrifier LSTM} & 35M
        & \nlltoppl{4.05079} & \nlltoppl{4.02249}
        & \nlltoppl{3.68904} & \nlltoppl{3.65259} \\
    & \textbf{Mogrifier RLSTM} & 35M
        & \nlltoppl{4.03738} & \nlltopplbold{4.00680}
        & \nlltoppl{3.67189} & \nlltopplbold{3.63711} \\
    \midrule
  \end{tabular}
\end{table*}

\section{Optimization}

To increase the stability of optimization and allow slightly higher learning rates, we use Rectified Adam \citep{liu2019variance}.
If training diverges, we reset the weights and the optimization state to the previous best checkpoint and multiply the learning rate by 0.9.

\citet{merity2017regularizing} switch to averaging weights at a late stage of optimization when the validation loss has not decreased for a while.
A similar procedure, called Stochastic Weight Averaging (SWA), was also proposed \citep{izmailov2018averaging}, and corresponding theory was developed in \citet{jain2018parallelizing} under the name of Tail Averaging.
Here, we employ Two-Tailed Averaging \citep{https://doi.org/10.48550/arxiv.2209.12581}, which has no hyperparameters and provides a good approximation to the optimal weight average at every step of optimization.
Two-Tailed Averaging (\tta{}) thus requires no tuning and is a much better fit with early stopping, which is what we do on \enwik and \texteight \citep{hutter2012human}.
While easier to work with, \tta{} does not improve the final results over well-tuned Tail Averaging, which is also used by our baseline, the Mogrifier LSTM.

Forget gates are initialized with Chrono init \citep{tallec2018can} as $\vb^f \sim \ln \mathcal{U}(1, T_\text{max}-1)$, where $T_\text{max}$ is a tuned hyperparameter.
Finally, we just train models longer where it is beneficial.

\section{Dynamic Evaluation}

\citet{hinton1987using} proposed fast weights, wherein parameters have a slow- and a fast-changing component.
Much later, \citet{ba2016using} showed a form of attention to be an instantiation of fast weights.
Similarly, some forms of meta learning, e.g.\ few-shot adaptation with gradient updates, can be interpreted as fast weights.
To mimic this setting and gain a particularly general fast weights implementation, it would be desirable to allow the model weights to depend on the context as in $p(x_i\tmid \theta(x_{<i}),x_{<i})$, where the fast weights $\theta(x_{<i})$ are computed with gradient-based updates to the slow weights $\theta_0$.
However, due to the practical difficulties involved in training batches with per-example inner gradient updates, we eschew fast weights at training time but not when performing evaluation, thus we end up with what is called dynamic evaluation \citep{krause2017dynamic}.
Here, we present dynamic evaluation results as a proxy for the gradient-based fast weights model.

As argued heuristically above, attention may be interpreted as a particular way of adapting the weights to the context (outer product memory), like the Mogrifier, which has an even more restricted form of update (scaling columns of weight matrices).
Thus, it is not surprising that \citet{melis2019mogrifier} found that Transformers are better at adapting without changing their weights, leaving less in-context signal for dynamic evaluation to pick up.

\begin{table*}
  \setlength{\abovecaptionskip}{0.2\baselineskip}
  \caption[Bits per character on character-based datasets of near state-of-the-art models.]{Bits per character on character-based datasets of near state-of-the-art models.
Names for models with our new results are in bold.
The best test results with and without dynamic evaluation for a given dataset and model size are in bold unless there is a smaller model with a better result.}
  \label{tab:cb-character-results}

  \centering
  \small
  \begin{tabular}{@{}llrllll@{}}
    & & & \multicolumn{2}{c}{No Dyneval} & \multicolumn{2}{c}{Dyneval} \\
    \cmidrule(lr){4-5} \cmidrule(lr){6-7}
    & & & Val. & \multicolumn{1}{r}{Test} & Val. & \multicolumn{1}{r}{Test} \\
    \midrule
    \parbox[t]{5mm}{\multirow{3}{*}{\rotatebox[origin=c]{90}{PTB}}}
    & Mogrifier LSTM \citep{melis2019mogrifier} & 24M
        & \nlltobpc{0.79616} & \nlltobpc{0.78415}
        & \nlltobpc{0.76119} & \nlltobpc{0.75439} \\
    & \textbf{Mogrifier LSTM} & 24M
        & \nlltobpc{0.78156} & \nlltobpc{0.76895}
        & \nlltobpc{0.75215} & \nlltobpc{0.74422} \\
    & \textbf{Mogrifier RLSTM} & 24M
        & \nlltobpc{0.77313} & \nlltobpcbold{0.75999}
        & \nlltobpc{0.74369} & \nlltobpcbold{0.73561} \\

    \midrule
    \parbox[t]{5mm}{\multirow{12}{*}{\rotatebox[origin=c]{90}{\enwik}}}
    & Transformer-XL (d24) \citep{dai2019transformer} & 277M
        & & 0.993 & & 0.940 \\
    & Longformer \citep{beltagy2020longformer} & 277M
        & & \textbf{0.97} & & \\
    \cmidrule(l){2-7}
    & Transformer-XL (d18) \citep{dai2019transformer} & 88M
        & & 1.03 & & \\
    & Longformer \citep{beltagy2020longformer} & 102M
        & & \textbf{0.99} & & \\
    & Mogrifier LSTM \citep{melis2019mogrifier} & 96M
        & \nlltobpc{0.76935} & \nlltobpc{0.77792}
        & \nlltobpc{0.69911} & \nlltobpc{0.68516} \\
    & \textbf{Mogrifier LSTM} & 96M
        & \nlltobpc{0.73297} & \nlltobpc{0.74346}
        & \nlltobpc{0.66784} & \nlltobpc{0.65592} \\
    & \textbf{Mogrifier RLSTM} & 96M
        & \nlltobpc{0.71288} & \nlltobpc{0.72236}
        & \nlltobpc{0.65992} & \nlltobpcbold{0.64803} \\
    \cmidrule(l){2-7}
    & Transformer-XL (d12) \citep{dai2019transformer} & 41M
        & & 1.06 & & 1.01 \\
    & Longformer \citep{beltagy2020longformer} & 41M
        & 1.02 & \textbf{1.00} & & \\
    & Mogrifier LSTM \citep{melis2019mogrifier}& 48M
        & \nlltobpc{0.78663} & \nlltobpc{0.79413}
        & \nlltobpc{0.71751} & \nlltobpc{0.70177} \\
    & \textbf{Mogrifier LSTM} & 48M
        & \nlltobpc{0.75051} & \nlltobpc{0.75813}
        & \nlltobpc{0.68495} & \nlltobpc{0.67213} \\
    & \textbf{Mogrifier RLSTM} & 48M
        & \nlltobpc{0.73502} & \nlltobpc{0.74228}
        & \nlltobpc{0.68362} & \nlltobpcbold{0.67110} \\

    \midrule
    \parbox[t]{5mm}{\multirow{6}{*}{\rotatebox[origin=c]{90}{\texteight}}}
    & Transformer-XL (d24) \citep{dai2019transformer} & 277M
        & & \textbf{1.08} & & \textbf{1.038} \\
    \cmidrule(l){2-7}
    & \textbf{Mogrifier LSTM} & 96M
        & \nlltobpc{0.71598} & \nlltobpc{0.76681}
        & \nlltobpc{0.67725} &  \nlltobpc{0.72559} \\
    & \textbf{Mogrifier RLSTM} & 96M
        & \nlltobpc{0.70855} & \nlltobpc{0.75983}
        & \nlltobpc{0.67595} & \nlltobpc{0.72363} \\
    \cmidrule(l){2-7}
    & Longformer \citep{beltagy2020longformer} & 41M
        & 1.04 & \textbf{1.10} & & \\
    & \textbf{Mogrifier LSTM} & 48M
        & \nlltobpc{0.73670} & \nlltobpc{0.79002}
        & \nlltobpc{0.69642} & \nlltobpc{0.74544} \\
    & \textbf{Mogrifier RLSTM} & 48M
        & \nlltobpc{0.72362} & \nlltobpc{0.77562}
        & \nlltobpc{0.69199} & \nlltobpc{0.74048} \\
    \midrule
  \end{tabular}
\end{table*}

\section{Experimental Setup}

We follow the experimental setup of our baseline \citep{melis2019mogrifier}.
In the following, we list only the most pertinent choices in our experimental setup; everything else is the same as in the baseline.
Note that the baseline's and our LSTM implementation already includes the capped input gate ($\min(1-\vf,\vi)$) of the RLSTM.
For the black-box hyperparameter tuner \citep{golovin2017google}, due to the switch to a residual architecture, the baseline's \emph{inter\_layer\_dropout} hyperparameter is replaced with \emph{cell\_output\_dropout} (see $\mM^{\textrm{cell},l}$ in \Cref{sec:architecture}).
In addition, the top of the Chrono init range $T_{\textit{max}}$ is a new hyperparameter from the $[e^2,e^5]$ range.

For word-level language modelling on Penn Treebank \citep{marcus1993building} with preprocessing by \citep{mikolov2010recurrent}, we trained 2-layer models for about 400 epochs with 8 dropout samples.
Batch size was 128, and we trained with a BPTT \citep{werbos1990backpropagation} window size of 70.
Experiments on \wikitexttwo \citep{DBLP:journals/corr/MerityXBS16} were conducted similarly, with the exception of training for 250 epochs and using 4 dropout samples.

For character-based language modelling on Penn Treebank, we trained 2-layer models for 100 epochs with 4 dropout samples, batch size 128, and a BPTT window size of 200.
On \enwik and \texteight \citep{hutter2012human}, we trained 6-layer models for 200 epochs, whereas the baseline model had 4 layers and was trained for only 29 epochs.
Batch size was 128, and the BPTT window size was set to 256.
Increasing the window size had no discernible effect in agreement with the findings of \citet{khandelwal2018sharp}.
Due to the long time required to train these models, the best hyperparameters were selected from a random pool of 60 candidates.
Since preliminary experiments indicated that the benefit of using multiple dropout samples was less than 0.01 bpc, we refrained from using more than one dropout sample to save time.

Model evaluation was performed with the standard, deterministic dropout; we refrained from using the more expensive Monte Carlo averaging \citep{gal2016theoretically}.
The optimal softmax temperature was selected at evaluation time to maximize the validation log-likelihood \citep{melis2018pushing}.
Finally, we report results with and without dynamic evaluation \citep{krause2017dynamic}.

\section{Results}

Due to the aforementioned loss of parallelization opportunities, we found that the RLSTM was about 10\%--30\% slower than the LSTM on NVIDIA p100 GPUs.
Still, the RLSTM retained a significant advantage over the LSTM even when trained for the same wall clock time.

On word-level language modelling (see \Cref{tab:cb-word-results}), only training for half the epochs, our Mogrifier LSTM outperformed the same model in \citet{melis2019mogrifier}, the baseline.
This was mostly due to the quicker convergence with multiple dropout samples and, to a small degree, to the initialization of the forget gate.
Note that on 2-layer models, which we used on this task, residual connections are identical to skip connections, which are employed by the baseline.
On top of these improvements, the RLSTM outperformed the LSTM by a small margin, and we established a new state of the art on both datasets with and without dynamic evaluation.

\Cref{tab:cb-character-results} shows our results on character-based language modelling.
On Penn Treebank, again only training for half the epochs, we significantly boosted the results of the state-of-the-art baseline model by using multiple dropout samples and to a smaller degree by tuning the forget gate initialization.
Our results were then further improved by switching to the RLSTM.

Our results on \enwik and \texteight were boosted greatly.
However, some corners were cut due the high computational cost, thus the results can likely be improved further e.g.\ with proper tuning (instead of random sampling), by training even longer, and by using multiple dropout samples.
We heuristically estimate a 0.015--0.03 bpc suboptimality due to these factors only.

Taking advantage of Two-Tailed Averaging's (\tta{}) online estimates and the fact that hyperparameters were selected randomly, we can estimate the contribution of training longer (200 vs 29 epochs).
For example, in the 48M parameter setting on \enwik, we observed that training longer lowered the validation bpc by about 0.035, leaving another 0.016 bpc for the contributions of residual connections and the forget gate initialization.
Finally, the RLSTM outperformed the LSTM by 0.022 bpc.
Similar relative contributions were observed in all other settings on \enwik and \texteight.

Some of the changes we made increased primarily the stability of training and the efficiency of hyperparameter tuning without discernible effect on the final results.
Using Rectified Adam and restarting from a previous checkpoint on divergence greatly reduced the number of failed runs, and \tta{} made tuning easier by removing a hyperparameter while opening the door for early stopping due to its online nature.

We found that in the early stages using $D$ dropout samples allowed optimization to make more progress per step but less than the progress with a single sample and $D$ times the number of steps.
On datasets that are on the smaller side and where dropout rates are high, in the late stages of optimization, the multi-sample objective proved better in terms of validation perplexity.
However, that was not the case on \enwik and \texteight, possibly due to being bottlenecked by other issues such as the length of optimization.

\section{Conclusions}

We strengthened purely recurrent models' language modelling results on a varied collection of datasets using various techniques and a slightly novel recurrent cell.
These new baselines better represent the models' ability but do not necessarily allow for a fair comparison to previous results.
In particular, while we cited and listed results of the best transformer-based models, we did not evaluate them ourselves using the same methodology.
This is especially obvious where previous works did not report dynamic evaluation results.
The comparisons we made to recurrent models were also rather targeted: we focussed solely on the state-of-the-art purely recurrent model, the Mogrifier LSTM, and ignored both the less performant models and those combined with attention \citep{merity2019single,DBLP:journals/corr/abs-2102-12459}.
Nevertheless, the combination of our strong results and little novelty highlight the extent to which minor details and computational efficiency on current hardware affect model comparisons.

With that in mind, the primary contribution of this work is to apply more resources to designing and evaluating recurrent models and to provide stronger baselines for model comparisons.
Second, the set of improvements to models of the recurrent cell, overall architecture, training objective, and optimization that we employed or proposed might inform future practice and research.
In addition, the fact that two quite different architectures (recurrent and attention-based) appear to perform similarly in terms of perplexity suggests that bigger changes are necessary to develop data-efficient language models.

{
  \ifdef{\groundskip}{}{\clearpage}
 %%   \interlinepenalty=10000
  \bibliography{paper}
  \bibliographystyle{plainnat}
  % KLUDGE: undo the last skip by natbib.
  \ifdef{\groundskip}
        {\vskip\lastskip
         \addvspace{2\groundskip}}
        {}
}

\ifdef{\groundskip}{}{\clearpage}

\begin{appendices}

\crefalias{section}{appendix}

\end{appendices}

\end{document}